\documentclass{article}
\usepackage[utf8]{inputenc}
\usepackage[english]{babel}
\usepackage{helvet}
\usepackage{algorithm}
\usepackage{algorithmic}
\usepackage{amsmath}
\usepackage{amssymb}

\usepackage[a4paper, total={5.9in, 9in}]{geometry}

\usepackage{graphicx}

\begin{document}

\thispagestyle{empty}
\begin{center}
    \large
%%%%%%%%%% Insert your title
    \textbf{Architectural Foundations for the Large Language Model Infrastructures}
    
%%%%%%%%%% Insert Author name      
    \vspace{0.4cm}
    \normalsize
    \textbf{Hongyin Zhu}
    
%%%%%%%%%% Insert Department         
    % \vspace{0.4cm}
    % \textit{Department address written in italic}
    
%%%%%%%%%% Insert your Email    
    \vspace{0.4cm}
    \textit{hongyin\_zhu@163.com}
    
    \vspace{0.4cm}
       
\end{center}
\begin{center}
\textbf{Abstract}
\end{center}
The development of a large language model (LLM) infrastructure is a pivotal undertaking in artificial intelligence. This paper explores the intricate landscape of LLM infrastructure, software, and data management. By analyzing these core components, we emphasize the pivotal considerations and safeguards crucial for successful LLM development. This work presents a concise synthesis of the challenges and strategies inherent in constructing a robust and effective LLM infrastructure, offering valuable insights for researchers and practitioners alike.
\begin{center}
\rule{0.9\linewidth}{0.4pt}
\end{center}
\vspace{0.2cm}

\noindent 
In the endeavor of constructing a robust and expansive large language model (LLM) infrastructure \cite{zhu2024climatechangelargelanguage}, the pivotal challenges encountered inherently encompass the trifecta of infrastructure, software, and data. These three fundamental components align seamlessly with the cornerstone pillars of computing power, algorithms, and data, respectively, forming the bedrock upon which the development of LLMs is anchored. The intricate interplay between these elements is paramount in fostering the advancement and refinement of artificial intelligence systems.

\section{Infrastructure Configuration} 

In the realm of infrastructure configuration for LLM training endeavors, server clusters equipped with H100/H800 GPUs have emerged as the de facto choice. Boasting exceptional computing prowess, these high-end GPUs offer a substantial reduction in training time by approximately 50\% in comparison to the A100 series, thereby facilitating a remarkable acceleration in model iteration and debugging processes. Notably, a cluster architecture comprising 8 nodes is sufficiently capable of completing the training process of a 7 billion parameter (7B) model within a single day, significantly narrowing the timeline from experimentation to deployment. Furthermore, the strategic selection and implementation of cluster management software is paramount, as they serve as the cornerstone for ensuring efficient resource allocation, scheduling, and the overall stability of cluster operations. While storage \cite{zhu2023heterogeneous} solutions may be economically feasible, the demands of LLM training \cite{zhu2023pre} necessitate vast amounts of storage capacity to accommodate the voluminous training data. Additionally, the networking infrastructure \cite{zhu2024node} is crucial for building a robust data center, enabling seamless data transfer and communication between various components of the system. Therefore, a comprehensive approach that addresses both storage and networking requirements is essential for ensuring the successful deployment and operation of LLM training infrastructures.

During the fine-tuning phase of LLMs, the advent of lightweight methods such as LoRA (Low-Rank Adaptation) has led to a notable reduction in computing power requirements. Consequently, GPUs from the A100/A800 series, as well as consumer-grade high-end GPUs like the RTX 4090/3090, are viable options for executing this task. While consumer-grade GPUs may exhibit slightly inferior fine-tuning efficiency, their inclusion underscores the versatility and accessibility of modern AI technologies.

In the realm of deploying LLM infernece systems, the escalating demand for computing power poses a formidable challenge. To effectively address this, precise estimations of the potential user base, coupled with profound software-level optimizations, are paramount in mitigating costs and enhancing deployment efficiency. For large-scale infernece tasks, a strategic selection of high-performance GPUs, such as the RTX 4090 and A100, offers a flexible solution. Additionally, the adaptability of resource allocation mechanisms is crucial, necessitating the dynamic adjustment of deployment strategies in response to actual demands, thereby striking a balance between computational requirements and cost-effectiveness. It is noteworthy that while GPUs undeniably excel in accelerating LLM inference, CPUs can likewise contribute as viable computing resources in select scenarios. In particular, when real-time constraints are less stringent and cost containment is a primary concern, leveraging multi-core CPUs for distributed inference deployment emerges as a viable alternative that merits further exploration. This approach underscores the importance of considering a diverse array of computational resources to cater to the unique requirements of various AI applications. The successful deployment of LLM inference necessitates a meticulous consideration of multifaceted factors, encompassing computational power prerequisites, cost-efficiency, software optimization strategies, and hardware selection. By adopting a scientific and well-conceived planning and execution framework, we can harness the full potential of these models, thereby fostering the ubiquitous adoption of AI applications and reinforcing their foundational support within various domains \cite{tiwari2021dapath}. This approach ensures that the deployment process is not only feasible but also sustainable, maximizing the impact and value of these advanced technologies.

\section{Software Framework}
In the meticulous design of software architecture, the choice and formulation of model frameworks assume a pivotal role. When navigating the choice between open-source and closed-source big model technology solutions, a delicate balance must be struck. Open-source models, with their virtues of transparency, scalability, and robust community support, offer researchers and developers a vast playground for exploration and innovation. Conversely, closed-source models, with their proprietary algorithm optimizations, performance advantages, and commercial backing, may emerge as the preferred choice for specific use cases. In making this crucial decision, it is imperative to delve deeply into one's own requirements, resource constraints, and long-term strategic vision. Only by doing so can one select the big model technology solution that aligns most closely with the unique demands and aspirations of the given scenario. With comprehensive support for a wide range of LLM resources \cite{zhu2023metaaid}, open-source frameworks enable developers to build on existing foundations and enhance the versatility, multimodality, and cross-domain generalization capabilities of their models. Furthermore, through the refinement and optimization of pre-training methodologies, these models are able to capture nuanced, personalized data characteristics, thereby laying a robust foundation for the seamless transition and application to subsequent tasks.

The advent of LoRA fine-tuning technology has emerged as a pivotal tool in optimizing model performance tailored to the unique demands of specific industries or business domains. This approach, characterized by its high efficiency and flexibility, leverages the incorporation of low-rank matrices to achieve the delicate balance of fine-tuning a minimal subset of parameters while preserving the vast majority of the original model's parameters unaltered. This strategic adaptation enables swift accommodation to novel tasks or domains, minimizing computational overhead while maintaining a substantial portion of the original model's knowledge base. Consequently, the performance and stability of the fine-tuned model are safeguarded, ensuring a seamless transition and optimized outcomes tailored to the specific application context.

Furthermore, the meticulous design of hyperparameters constitute a pivotal juncture in enhancing model performance. Serving as a vital interface between the model's architecture and its learning dynamics, the judicious configuration of hyperparameters is imperative for optimizing model efficacy. Consequently, it is imperative to embark on a methodical exploration, grounded in a rigorous experimental design and validation framework, aimed at identifying the optimal hyperparameter configuration tailored to specific tasks. This endeavor necessitates a harmonious fusion of domain expertise, engineering practices, and a robust theoretical foundation. Researchers must possess not only a profound understanding of the underlying principles but also a wealth of practical experience and a keen intuition to navigate this intricate landscape.

The alignment mechanism of big models, a cornerstone technology ensuring adherence to compliance and ethical standards amidst their pervasive application, underpins the legitimacy, fairness, and reliability of model decisions, outputs, and behaviors. This multifaceted approach encompasses strategies such as data compliance validation, augmentation of model transparency, bias detection methodologies, and robust privacy protection measures. To establish a credible, reliable, and responsible AI system, this process necessitates rigorous screening of training data, intensification of model interpretability, scrutiny of model outcomes, and exhaustive pre-deployment testing. By adhering to these principles, we can foster the healthy and orderly evolution of big model technology, ultimately contributing positive societal impacts and advancing the field of artificial intelligence.

In the context of LLM deployment, the significance of algorithm optimization cannot be overstated. The transition from research and development (R\&D) to production environments poses myriad challenges, including constraints on computing resources, stringent real-time requirements, and paramount concerns regarding data security and privacy protection. To ensure efficient, stable, and secure operation of the model in practical applications, it is imperative to devise and implement tailored algorithm optimization strategies, grounded in the unique characteristics of the deployment environment. These strategies encompass a broad spectrum of techniques, including but not limited to model pruning, quantization, knowledge distillation, and distributed inference, each tailored to address specific challenges. Furthermore, leveraging specialized reasoning frameworks and libraries, such as LMDeploy, vLLM, Ollama, and llama.cpp, optimized for hardware acceleration, can significantly enhance performance. The comprehensive application of these technologies effectively mitigates the computational burden and memory footprint during inference, thereby reducing reliance on robust hardware resources while maintaining the model's predictive capabilities.

In the realm of big model applications, the front-end presentation layer offers a versatile palette of modern frameworks, such as Streamlit and Gradio, alongside traditional HTML-based front-end and back-end separation technologies. These solutions provide a flexible and diverse canvas for crafting user interfaces that cater to diverse needs. Additionally, by offering independent API services, big model technology transcends traditional constraints, seamlessly integrating into a myriad of applications and service platforms, fostering seamless data flow and facilitating in-depth value extraction. The advent of solutions that support end-side deployment further expands the horizons of big model applications, enabling them to operate agilely on various terminals, including edge devices and mobile devices. This capability underscores the potential for delivering more secure, reliable, and personalized services to users. 

\section{Data Management}

The cultivation of data at a profound level necessitates the implementation of efficient and rigorous data management strategies, which serve as the bedrock for the success of big models. This endeavor encompasses a multifaceted approach that prioritizes data integrity verification, ensuring that every data point originates from a trustworthy source and remains unadulterated. Furthermore, it encompasses pivotal steps such as the adjustment of data category balance, the meticulous filtering of noise, and the diligent detection of duplications. These measures, collectively, contribute to the construction of a high-quality dataset, thereby laying a robust foundation for the subsequent training and optimization of models. By adhering to these principles, researchers and practitioners can ensure that their LLMs are grounded in data that is both reliable and representative.

To maximize the efficacy of model learning, the application of refined data engineering has become increasingly imperative. This engineering methodology involves the meticulous processing and refinement of raw data through a series of meticulously designed processes and technical methodologies, with the ultimate goal of enhancing the intrinsic quality and representativeness of the data. Specifically, it necessitates the scientific planning of sample proportions within each category within the dataset, ensuring that the model can learn the characteristics of each category in a balanced and unbiased manner during the training process. Additionally, the utilization of noise filtering techniques is crucial for eliminating invalid or interfering information within the data, thereby mitigating its negative impact on model learning. Furthermore, the detection of duplicate samples is an indispensable component of this process, as it aims to eliminate redundancies and prevent overfitting issues during model training. By adopting these refined data engineering practices, researchers and practitioners can foster the development of more robust and accurate AI systems.

The formulation of a data matching plan necessitates a meticulous focus on the unique characteristics of the task at hand and the specific requirements of the target application. This involves a thorough analysis of the influence that various types of data have on model performance, followed by the development of a matching strategy that comprehensively addresses the task requirements while precisely capturing the pertinent information within the data. This endeavor necessitates not only a profound understanding of the relevant domain but also the proficient utilization of advanced data analysis tools and technical methodologies to achieve optimal data matching. By adhering to these principles, researchers and practitioners can enhance the accuracy and effectiveness of their AI models, thereby fostering the development of more sophisticated and reliable AI systems.

\section{Conclusion}
This paper delves into the multifaceted considerations that are paramount when constructing a robust big model infrastructure. Central to this endeavor is the computational prowess of the infrastructure, which serves as the backbone for the demanding computations required by big models. Furthermore, the flexibility and scalability of the underlying software architecture are essential for ensuring that the infrastructure can adapt to the evolving needs of big models and facilitate their seamless integration into diverse application domains. Additionally, the abundance and quality of data resources are crucial factors, as they provide the fuel that drives the learning and innovation of big models. The complementary nature of these factors – computational power, software architecture, and data resources – collaboratively fosters the innovative application and widespread implementation of big model technology across a myriad of fields within the realm of AI.

% \begin{figure}[h]
%     \centering
%     \includegraphics[width=0.65\textwidth]{DuckPicture.pdf}
%     \caption{Figure captions should be below the figure.}
%     \label{fig:my_label}
% \end{figure}

% \printbibliography
\bibliographystyle{unsrt} % 指定参考文献的样式  
\bibliography{Literature} % 指定.bib文件名，不需要.bib扩展名 
\end{document}